\def\year{2022}\relax
\documentclass[letterpaper]{article} 
\usepackage{aaai22}  
\usepackage{times}  
\usepackage{helvet}  
\usepackage{courier}  
\usepackage[hyphens]{url}  
\usepackage{graphicx} 
\usepackage{natbib}  
\usepackage{caption} 
\usepackage{balance}
\DeclareCaptionStyle{ruled}{labelfont=normalfont,labelsep=colon,strut=off} 
\usepackage{algorithm}
\usepackage{algorithmic}
\usepackage[dvipsnames]{xcolor}
\usepackage{newfloat}
\usepackage{listings}
\floatstyle{ruled}
\newfloat{listing}{tb}{lst}{}
\floatname{listing}{Listing}
\usepackage{amsmath}
\usepackage{amssymb}
\usepackage{amsfonts}
\usepackage{bbm}

\title{Interpret Your Care: 
Predicting the Evolution of Symptoms for Cancer Patients
}
\author{
    Rupali Bhati \textsuperscript{\rm 1},
    Jennifer Jones \textsuperscript{\rm 2},
    Audrey Durand \textsuperscript{\rm 1,3} 
}
\affiliations{

    \textsuperscript{\rm 1} Université Laval \\
    \textsuperscript{\rm 2} Princess Margaret Cancer Centre, University Health Network \\
    \textsuperscript{\rm 3} CIFAR AI Chair

}

\usepackage{bibentry}
\begin{document}

\maketitle

\begin{abstract}

Cancer treatment is an arduous process for patients and causes many side-effects during and post-treatment. The treatment can affect almost all body systems and result in pain, fatigue, sleep disturbances, cognitive impairments, etc. These conditions are often under-diagnosed or under-treated. In this paper, we use patient data to predict the evolution of their symptoms such that treatment-related impairments can be prevented or effects meaningfully ameliorated. The focus of this study is on predicting the pain and tiredness level of a patient post their diagnosis. We implement an interpretable decision tree based model called LightGBM on real-world patient data consisting of 20163 patients. There exists a class imbalance problem in the dataset which we resolve using the oversampling technique of SMOTE. Our empirical results show that the value of the previous level of a symptom is a key indicator for prediction and the weighted average deviation in prediction of pain level is 3.52 and of tiredness level is 2.27.

\end{abstract}

\section{Introduction}
Nearly half of Canadians are expected to receive a diagnosis of cancer in their lifetime~\cite{BrennerE199}. 
Unlike other chronic diseases 
where disability is commonly caused by the disease process, 
disability associated with cancer is often caused more by treatment than the disease itself~\cite{short2008work}. 
Individual cancer patients may undergo different treatments, for different amounts of time and may even
suffer from multiple diagnoses. Such experiences render the patients with distinct symptoms along their treatment, for example pain, fatigue, sleep disturbances, or cognitive impairments. 
There is a need for tailored models of 
rehabilitation that account for the fact that “one size is not likely to fit all” needs of different cancer survivors \cite{raymond2021future}. 
To this aim, in many cancer centres, patient symptom data is collected via surveys which are filled by the patient when they come for their cancer clinic visits. 
However, asking patients to fill in surveys about their health conditions too often can be very burdening. 
Therefore, there is a need to systematically identify the adverse effects of cancer such that they can be treated efficiently.

Our aim is to leverage previously collected patient reported data 
to 
predict the evolution of a patient's symptoms.
Predicting when a patient encounters high levels of symptom burden may help clinicians take precautionary 
measures to avoid such situations or be able to effectively manage the patient's conditions. 
For a predictive model to be used within such an application, it should be interpretable by clinicians~\cite{zhu2014using}. The clinician not only needs to see the predicted value of the symptom but also needs to understand the criteria based on which the prediction is made.

In this paper we, therefore, propose an interpretable pipeline for predicting the level of two different symptoms, i.e. pain and tiredness levels, using LightGBM~\cite{NIPS2017_6449f44a}, a decision tree (DT) based model which is widely used for its efficiency and interpretability. DT based methods have a wide application in healthcare~\cite{Podgorelec2002decision,diagnostics11091714} and cancer-related applications, for example to predict breast cancer survivability~\cite{4650373}.
Although other, possibly more complex, predictive models have displayed good performances at predicting future symptoms of oncology patients~\cite{papachristou2018learn},
there has been resistance to adopting these machine learning based methods in clinical practice due to a perception that these are “black-box” techniques and incompatible with decision-making based on evidence and clinical experience~\cite{zhu2014using}. Therefore, the proposed pipeline rather relies on an interpretable approach based on DTs, combined with a training procedure based on oversampling~\cite{chawla2002smote} to compensate for the important data imbalance observed across the different values of symptom level. By evaluating this approach against a naive predictor based solely on prior distributions of symptoms, we highlight the necessity of using evaluation metrics that compensate for data imbalance in order to evaluate fairness in model performances. Finally, although a feature importance analysis of our models reveals that the previous symptom value plays an important role in predicting the current symptom value, our results show a clear benefit in leveraging clinical variables in the predictions and, in some cases, the correlation between symptoms.

\section{Problem Statement}
\label{sec:problem_statement}

\begin{figure}[tb]
\centering
\includegraphics[width=0.45\textwidth]{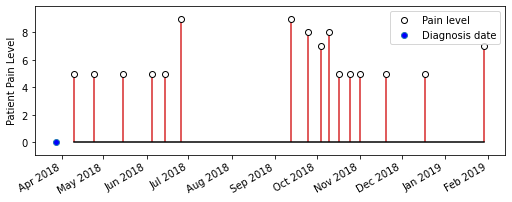}
\includegraphics[width=0.45\textwidth]{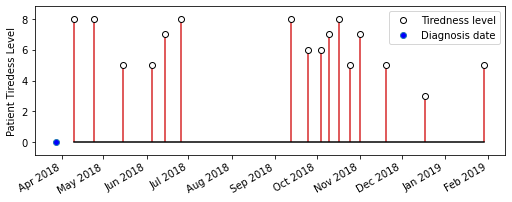}
\caption{Example of symptom evolution for one patient}
\label{fig:Patient_trajectories}
\end{figure}

Post diagnosis, a patient visits the doctor or clinician either when they are going through a treatment or require some assistance. These visits can take place at irregular intervals. During such visits, the patient is asked to fill a survey documenting their symptoms. 
Pain and tiredness are two symptoms which acutely affect the overall quality of life of a patient undergoing cancer treatment.
Both symptoms are documented by the patient as a discrete value between 0 and 10, where 0 is the lowest level of pain/tiredness a patient can feel and 10 is the highest based on the Edmonton Symptom Assessment Scale (ESAS)~\cite{watanabe2012edmonton}. 
To demonstrate how the symptoms of a patient can evolve over a period of time,  Figure~\ref{fig:Patient_trajectories} shows an example of the evolution of a patient's symptoms along a timeline. 

We have access to a dataset of real cases collected for
years
2013-2019 on 20163 patients. On average, there are 5.95 observations per patient.
%
Four cancer types are included in this analysis, namely, breast, colorectal, lymphoma and head and neck. 
The dataset includes patient information, i.e., sex, age, cancer type, and surgery data, in addition to the survey data, i.e., information about the patient's symptoms, including ESAS values and Patient Reported Functional Status (PRFS) measures~\cite{popovic2018patient} which monitor overall nutritional status in patients. 
For the purpose of this study, we focus on predicting some important symptoms using clinical and demographic information of the patient and their previous symptom values which are summarised in Table~\ref{tab:Inputs} along with the values each feature can take.






\begin{figure}[tb]
\centering
\includegraphics[width=0.47\textwidth]{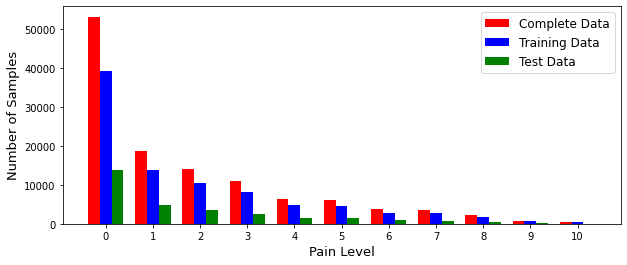}
\includegraphics[width=0.47\textwidth]{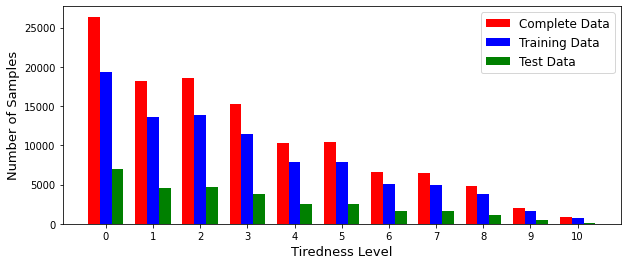}
\caption{Distribution of number of samples over classes for pain and tiredness levels}
\label{fig:DistributionofESASPainandTirednessLevels}
\end{figure}


To avoid a situation where the future observations of a patient are in the training set while the older observations are in the test set, we split the dataset based on a date. The surveys before the date '2017-10-04' are in the training set and the ones after this date are in the test set. This leads to a 75\%-25\% split between the training and test set, respectively. The distribution of number of samples in each class can been seen in Figure~\ref{fig:DistributionofESASPainandTirednessLevels} for each symptom.
We observe that there are class imbalance problems, where $\sim44\%$ and $\sim22\%$ of the outputs belong to one class (level 0) for pain and tiredness, respectively.
It is also noted that the number of samples in a class are inversely proportional to the symptom level.
This occurs because high levels of pain and tiredness levels are relatively rarer events. 
This bias in the data can lead to unfairness in the model's predictions and therefore inequity in a patient's treatment and may exacerbate health care disparities~\cite{Rajkomar2018ensuring}.





\begin{table}[tb]
 \begin{tabular}{|p{4.6cm}|p{3.2cm}|} 
 \hline
 \textbf{Variable} & \textbf{Possible values}\\
 \hline
 Sex &  male, female \\
 \hline
 Age & 18 to 93 \\
 \hline
 Cancer type  & breast, head and neck, lymphoma, colorectal  \\
 \hline
 Number of days between the diagnosis and the survey date
 & 5 to 23091 ($\sim63$ years)\\
 \hline
 Number of days between the previous and the current survey date & 1 to 2002 ($\sim5$ years)\\
 \hline
 Previous pain level
 &\{0,1,2,3,4,5,6,7,8,9,10\} \\
 \hline
 Previous tiredness level
 &\{0,1,2,3,4,5,6,7,8,9,10\} \\
 \hline
\end{tabular}
\captionof{table}{Variables available for prediction}
\label{tab:Inputs}
\end{table}



\section{Methodology}
\label{sec:methodology}

In this section, we propose a pipeline to predict the symptom levels of a patient while trying to maintain equal performance as much as possible over the different classes. As the values to predict are discrete (they belong to the finite, discrete set of symptom levels), we formulate this problem as a classification problem. We propose to evaluate candidate strategies using an interpretable metric that accounts for class imbalance, ensuring that the performance over all groups of patients (based on their symptoms) is equally represented in the global evaluation score.

\subsection{Predictive model}

Clinical providers and other decision makers in healthcare note interpretability of model predictions as a priority for implementation and utilization \cite{McKelvey2018Interpretable}. 
To this end, we use DTs for prediction as each decision point can be visualised individually by clinicians. Among several DT algorithms, we select 
LightGBM for several reasons.
Firstly, LightGBM is a gradient boosting DT algorithm. 
Gradient boosting is an approach where new models are created that predict the residuals or errors of prior models and then added together to make the final prediction. It is called gradient boosting because it uses a gradient descent algorithm to minimize the loss when adding new models which is more beneficial as compared to other DT algorithms that either train the different models separately before aggregating them or do not use multiple models at all.
Secondly, 
instead of building trees that have a uniform depth across all branches, LightGBM expands the tree leaf-wise (up to a maximum tree depth given as a parameter), resulting in  much lighter trees and making it suitable for high-speed and low memory consumption. 

For predicting each symptom (pain and tiredness), we consider two variants of LightGBM models. The first variant relies on clinical variables (first five rows of Table~\ref{tab:Inputs}) and previous level of the predicted symptom (row 6 or 7 of Table~\ref{tab:Inputs}) in the prediction, while the second variant uses all clinical variables in addition to previous levels of both symptoms (all rows of Table~\ref{tab:Inputs}). We refer to the first variants as LP1 and LT1, and to the second variants as LP2 and LT2, for pain and tiredness respectively.


\paragraph{Fighting data imbalance}

Inspired by previous works~\cite{ Zhao2018framework}, we rely on Synthetic Minority Oversampling TEchnique (SMOTE)~\cite{chawla2002smote} to address class imbalance by generating synthetic samples. At training time, SMOTE over-samples the minority classes so that the trained model sees the same number of data points of these classes as for the majority class. A minority class is over-sampled by taking each of its samples and introducing synthetic examples along the line segments joining any/all of the $k$ minority class nearest neighbours. 
For our experiments, we use
$k=5$.

\paragraph{Training}

To determine the maximum tree depth parameter for LightGBM, we perform a 5-fold cross-validation on the training set (90046 samples from 16674 patients) for tree depths values between 1 and 25 (as the performance saturates after 25). The maximum tree depth value minimizing the WMAE metric (described below using Equations~\ref{eqn:MAE},~\ref{eqn:CW},~\ref{eqn:WMAE}) averaged over the 5 folds is then used for retraining a LightGBM model using on the whole training dataset.

\subsection{Baselines}
We compare the performance of the LightGBM models with two baselines models. We refer to these baselines as NP and PV for naive prior and previous value, respectively.

\paragraph{Naive prior} This baseline is a naive strategy that always predicts the dominant (most frequent) symptom level (i.e., 0 for both pain and tiredness).

\paragraph{Previous value} This baseline is slightly smarter than NP. It predicts the current symptom level to be the same as the previous symptom level, which could have been recorded at any time in the past because surveys are collected at irregular intervals. Therefore, the previous value may or may not be a good indicator of the current value of the symptom.


\subsection{Metrics}


In case of imbalanced classes, there can often be misleading evaluation metrics where mistakes on a minority class can be hidden by good performance on a dominant class. This can lead to unfairness in the treatment of patients, for example, due to the high frequency of low pain level in the data, we might underestimate the level of pain a patient might be in.
To this end, we use weighted metrics such that the model can focus on achieving good performance on individual classes as well as overall.
Let $\mathcal C$ denote the set of classes. We measure the performance of a given strategy on each class $c\in \mathcal C$ using the mean absolute error (MAE) :
\begin{equation}
\label{eqn:MAE}
\text{MAE}_c = \frac{1}{N_{c}} \sum_{i=1}^{N_c} \left| y_i - \hat y_i \right|,
\end{equation}
where $y_i$ is the true value of the $i^{th}$ sample in class $c$, $\hat{y_i}$ is the value predicted by the strategy for the $i^{th}$ sample in class $c$, and $N_c$ is the number of samples belonging to class $c$. The MAE corresponds to the absolute deviation in the symptom level making it interpretable for clinicians to infer and analyse.
Therefore, a good model would have a low MAE so that for the correct predictions, it would be zero and for the incorrect predictions, the model would predict a symptom level closer to the true symptom level. We combine these values using the weighted-MAE (WMAE), which essentially gives value to the MAE of each class inversely proportional to the number of samples in that class. Therefore, mistakes on a class for which there is little data account for more in the WMAE.
To calculate the WMAE, we first calculate the weight of each class $c$ as
\begin{equation}
w_c = \frac{\max_{c'\in\mathcal C} N_{c'}}{N_{c}}.
\label{eqn:CW}
\end{equation}
The weight will therefore be of value 1 for the class with the largest amount of data, and would be 10 times larger for a class with 10 times less data.
%
We can then compute the global WMAE as
\begin{equation}
\text{WMAE}= \frac{\sum_{c\in\mathcal C} w_c \text{MAE}_c}{\sum_{c\in\mathcal C} w_c}.
\label{eqn:WMAE}
\end{equation}



\section{Results}
\label{sec:results}


\begin{table}[tb]
\centering
 \begin{tabular}{|p{1.1cm}|p{0.7cm}|p{0.9cm}|p{1.1cm}|p{1.3cm}|} 
 \hline
 & NP & PV & LP1 & LP2\\
 \hline
 \hline
 $\text{MAE}_0$ 
  & 0 & 0.52 & 0.8 & 0.8  \\
 \hline
 $\text{MAE}_1$ 
  & 1  & 0.91  & 1.34  &1.35\\
 \hline
 $\text{MAE}_2$ 
 & 2& 1.3  & 1.89  &1.88\\
 \hline
 $\text{MAE}_3$  
 &3& 1.5   & 2.11  & 2.09  \\
 \hline
 $\text{MAE}_4$ 
 &4& 1.82  & 2.3   &2.26  \\
 \hline
 $\text{MAE}_5$ 
 &5& 2.04   & 2.45 &  2.46  \\
 \hline
 $\text{MAE}_6$ 
 &6& 2.39 &   2.54  & 2.55  \\
 \hline
 $\text{MAE}_7$ 
 &7&  2.76   & 2.67   & 2.72   \\
 \hline
 $\text{MAE}_8$ 
 &8& 2.89  &  2.65  &  2.62     \\
 \hline
 $\text{MAE}_9$ 
 &9& 2.98  & 2.59 & 2.64    \\
 \hline
 $\text{MAE}_{10}$ 
 &10& 5.03 &  4.41 & 4.53   \\
 \hline
 \hline
 WMAE &  8.72 & 3.88 & \textbf{3.52} & 3.57   \\
 \hline
\end{tabular}
\caption{Prediction performance per pain level ($\text{MAE}_c$) and global performance (WMAE) for each strategy}
\label{tab:Results_Pain}
\end{table}


\begin{table}[tb]
\centering
 \begin{tabular}{|p{1.1cm}|p{0.7cm}|p{0.9cm}|p{1.1cm}|p{1.3cm}|} 
 \hline
 & NP & PV & LT1 & LT2\\
 \hline
 \hline
 $\text{MAE}_0$ 
   & 0 & 0.7  & 0.84  &  0.83 \\
 \hline
 $\text{MAE}_1$ 
   & 1 & 0.99  & 1.21  &1.21 \\
 \hline
 $\text{MAE}_2$ 
 &2& 1.22  & 1.6  & 1.59 \\
 \hline
 $\text{MAE}_3$  
 &3& 1.41  & 1.86  & 1.86  \\
 \hline
 $\text{MAE}_4$ 
 &4& 1.55  & 2.02 & 2.01 \\
 \hline
 $\text{MAE}_5$ 
 &5& 1.71  & 2.23  &  2.22   \\
 \hline
 $\text{MAE}_6$ 
 &6& 1.91  & 2.29  & 2.27  \\
 \hline
 $\text{MAE}_7$ 
 &7& 2.06 & 2.17 & 2.18  \\
 \hline
 $\text{MAE}_8$ 
 &8& 2.2  & 2.17   &  2.15      \\
 \hline
 $\text{MAE}_9$ 
 &9& 2.4  & 2 &   2.03    \\
 \hline
 $\text{MAE}_{10}$ 
 &10 & 3.25 & 2.63 &   2.52 \\
 \hline
 \hline
 WMAE &  8.43  & 2.65 &  2.33  & \textbf{2.27}  \\
 \hline
\end{tabular}
\caption{Prediction performance per tiredness level ($\text{MAE}_c$) and global performance (WMAE) for each strategy}
\label{tab:Results_Tiredness}
\end{table}

For each symptom, we evaluate the performance of the four considered strategies: the two baselines (NP and PV) and the two variants of LightGBM models (LP1/LT1 and LP2/LT2). We consider both the performance per class ($\text{MAE}_c$) and the global performance (WMAE).

Table~\ref{tab:Results_Pain} shows the results obtained for the task of pain level prediction. Recall that the dominant class for pain level is class 0, therefore, the NP model predicts 0 everywhere.
If we did not account for class imbalance, the average of the $\text{MAE}_c$'s would be 55/11 = 5, which is lower than the WMAE. Even worse, the global MAE would be 1.65. This shows how low MAE values for high frequency classes can compensate for the high MAE values of low frequency classes, ultimately leading to an unfair metric to evaluate the performance of a model. In that case, the model would predict high frequency classes more often and still be able to maintain a good average MAE. In application, the model would predict high levels of pain less often which would be detrimental to the patients who might need immediate care. The LP1 model performs the best overall suggesting that using the previous tiredness level is not 
useful to
predict the pain level. Although the discrepancy between class performance is slightly attenuated with the LP1 model compared with other strategies, we should still note that, its performance ($\text{MAE}_c$) is still inversely proportional to the class density. This indicates that the LP1 model is still unfair towards individuals with stronger pain and further work would probably be needed to make it clinically acceptable.

Table~\ref{tab:Results_Tiredness} shows the results obtained for the 4 models used to predict the tiredness level. The dominant class here is once again class 0. Unsurprisingly, we observe lower WMAE when predicting tiredness than pain since the former suffers less class imbalance.
Moreover, it can be observed that LT2 performs better than LT1 indicated by the lower WMAE. This suggests that adding the previous pain level is a good indicator to predict the tiredness level.

It is interesting to note that the advantage of LightGBM-based models is observed for higher symptom levels, i.e., $\text{MAE}_c$ for $c > 6$ is higher with PV than with LP1/LP2 and $\text{MAE}_c$ for $c > 7$ is higher with PV than with LT1/LT2. This suggests that patients with higher symptom levels would benefit the most from more advanced models.

\begin{figure}[tb]
\centering
\includegraphics[width=0.45\textwidth]{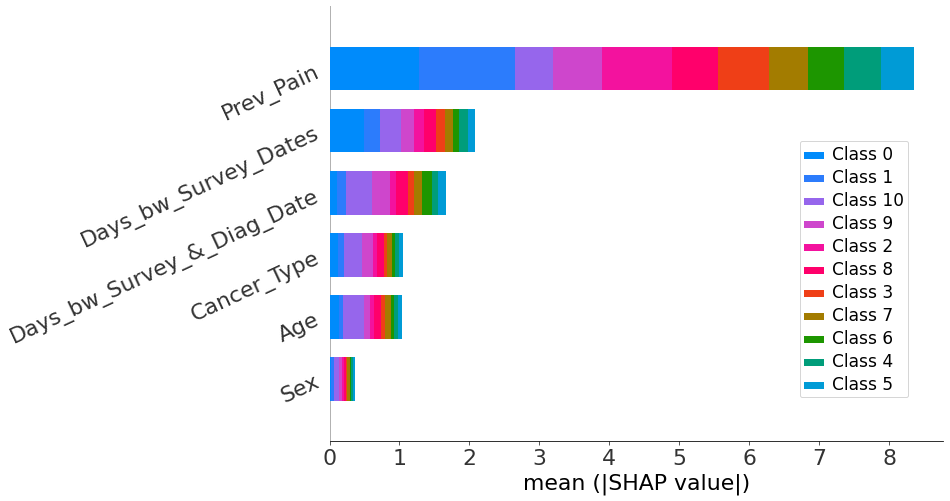}
\caption{Feature Importance obtained from LightGBM model to predict pain (LP1) }
\label{fig:ESASPainLGBMC}
\end{figure}

\begin{figure}[tb]
\centering
\includegraphics[width=0.45\textwidth]{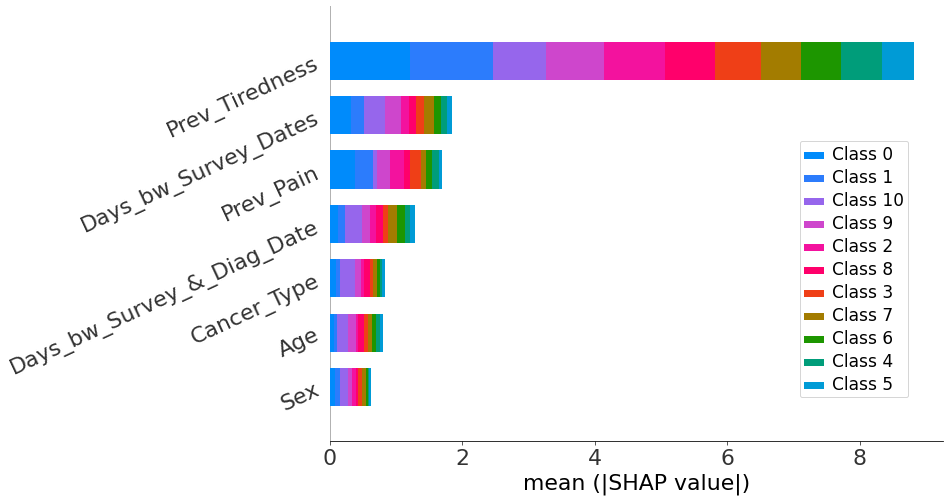}
\caption{Feature Importance obtained from LightGBM model to predict tiredness using previous pain level (LT2) }
\label{fig:ESASTirednessLGBMC}
\end{figure}

Figures~\ref{fig:ESASPainLGBMC} and~\ref{fig:ESASTirednessLGBMC} show the importance of each feature in terms of their SHAP~\cite{lundberg2017unified} values for the LP1 and LT2 models, respectively. The SHAP value of a feature indicates the impact of the feature on the magnitude of the output, helping to understand the contribution of each feature to each class. For both symptoms, we observe that previous symptom levels strongly contribute to the prediction for current symptom levels. 


\section{Conclusion}

In this study, we predict the pain and tiredness level of a patient using their clinical and demographic data and previous observations of the symptom
levels. The results demonstrate the need for weighted metrics and the importance of previous symptom levels. 
In the future, we want to formulate this problem as a  a reinforcement learning problem and model the context-reward relationship using the DTs as discussed in \cite{elmachtoub2018practical}. We also plan to predict more qualitative measures like the indicator of quality of life.

\section{Acknowledgements}

We thank Mathieu Godbout for his feedback on the paper and the CanRehab team for their generous support.

\balance
\bibliography{aaai22}

\end{document}